\documentclass{article} 
\usepackage[dvips]{graphicx}
\usepackage{amssymb,amsmath,color}
\usepackage{url}

\title{Maximizing submodular functions \\
using probabilistic graphical models}

\author{K. S. Sesh Kumar\\
INRIA-Sierra project-team\\ 
D{\'e}partement d{'}Informatique \\
de l{'}Ecole Normale Sup{\'e}rieure\\
Paris, France\\
\texttt{sesh-kumar.karri@inria.fr}\\
\and Francis Bach\\
INRIA-Sierra project-team\\ 
D{\'e}partement d{'}Informatique \\
de l{'}Ecole Normale Sup{\'e}rieure\\
Paris, France\\
\texttt{francis.bach@inria.fr}
}

 \newcommand{\defeq}{\stackrel{\rm def}{=}}

\newcommand{\BEAS}{\begin{eqnarray*}}
\newcommand{\EEAS}{\end{eqnarray*}}
\newcommand{\BEA}{\begin{eqnarray}}
\newcommand{\EEA}{\end{eqnarray}}
\newcommand{\BEQ}{\begin{equation}}
\newcommand{\EEQ}{\end{equation}}
\newcommand{\BIT}{\begin{itemize}}
\newcommand{\EIT}{\end{itemize}}
\newcommand{\BNUM}{\begin{enumerate}}
\newcommand{\ENUM}{\end{enumerate}}
\newcommand{\BA}{\begin{array}}
\newcommand{\EA}{\end{array}}

\newcommand{\rb}{\mathbb{R}}
\newcommand{\BlackBox}{\rule{1.5ex}{1.5ex}}  
\newcommand{\lova}{Lov\'asz }

\newtheorem{proposition}{Proposition}

\newcommand{\mysec}[1]{Section~\ref{sec:#1}}
\newcommand{\eq}[1]{Eq.~(\ref{eq:#1})}
\newcommand{\myfig}[1]{Figure~\ref{fig:#1}}
\def \E { \mathbb{E}}

\begin{document}
\maketitle

\begin{abstract}
We consider the problem of maximizing submodular functions; while this problem is known to be NP-hard, several numerically efficient local search techniques with approximation guarantees are available. In this paper, we propose a novel convex relaxation which is based on the relationship between submodular functions, entropies and probabilistic graphical models. In a graphical model, the entropy of the joint distribution decomposes as a sum of marginal entropies of subsets of variables; moreover, for any distribution, the entropy of the closest distribution factorizing in the graphical model provides an  bound on the entropy. For directed graphical models, this last property turns out to be a direct consequence of the submodularity of the entropy function, and allows the generalization of graphical-model-based upper bounds to any submodular functions. These upper bounds may then be jointly maximized with respect to a set, while minimized with respect to the graph, leading to a convex variational inference scheme for maximizing submodular functions, based on outer approximations of the marginal polytope  and maximum likelihood bounded
treewidth structures. By considering graphs of increasing treewidths, we may then explore the trade-off between computational complexity and tightness of the relaxation. We also present extensions to constrained problems and maximizing the difference of submodular functions, which include all possible set functions.
\end{abstract}

\section{Introduction}
\label{sec:intro}

Optimizing submodular functions has been an active area of research with applications 
in graph-cut-based image segmentation~\cite{boykov2001fast}, sensor placement~\cite{krause11submodularity}, or document summarization~\cite{lin2011-class-submod-sum}. 
A set function~$F$ is a function defined on the power set $2^V$ of a certain set $V$. It is submodular if and only if for all 
$A, B \subseteq V$, $F(A)+F(B)\geqslant F(A\cap B)+F(A\cup B)$.  Equivalently, these functions also admit the diminishing returns 
property, i.e., the marginal cost of an element in the context of a smaller set is more than its cost in the context of a larger 
set. Classical examples of such functions are entropy, mutual information, cut functions, and covering functions---see further 
examples in~\cite{fujishige2005submodular,submodtutBach}.

Submodular functions form an interesting class of discrete functions  because minimizing a submodular function can be done in polynomial 
time~\cite{fujishige2005submodular}, while maximization, although NP-hard, admits constant factor approximation algorithms~\cite{nemhauser1978analysis}. 
In this paper, our ultimate goal is to provide the first (to the best of our knowledge) generic convex relaxation of submodular function maximization, 
with a hierarchy of complexities related to known combinatorial hierarchies such as the Sherali-Adams hierarchy~\cite{SheraliA90}. Beyond the 
graphical model tools that we are going to develop, having convex relaxations may be interesting for several reasons: (1) they can lead to better solutions, 
(2) they provide online bounds that may be used within branch-and-bound optimization and (3) they ease the use of such combinatorial optimization problems 
within structured prediction framework~\cite{Tsochantaridis04}. 

Feige et al.~\cite{Feige2011} proposed constant factor approximation algorithms for maximizing non-negative submodular functions. They provide a randomized local search technique which 
optimizes a multilinear auxiliary function with some approximation guarantees. 
Buchbinder et al.~\cite{Buchbinder2012} proposed a ramdomized $1/2$-approximation algorithm to maximize non-negative submodular functions. They also use a randomized local search to remove
or add an element for the existing set under consideration in each iteration of the algorithm. 
However, these methods only consider unconstrained submodular maximization. 

Recent works also consider maximization of non-negative submodular functions~\cite{Vondrak2011} with packing-type constraints such as knapsack constraints, matroid constraints and 
their intersections with $0.309$-approximation guarentee with respect to the best integer solution on the matroid polytope. They consider an extreme point of the polytope and provide a technique to 
replace an element of the extreme point fractionally using linear optimization. Iyer et al.~\cite{iyer2013} proposed
semi-differentials, discrete equivalent of gradients, to define linear bounds on submodular functions. The approximations thus obtained are optimized using CCCP-like~\cite{yuille2003} procedures.

Among submodular functions, entropies have been particularly well-studied. Given $V = \{1, 2, \ldots, n\}$, we consider $n$ random variables 
$X_1, \ldots, X_n$ (jointly referred to as~$X$) where $\mathcal{X}=\mathcal{X}_1 \times \mathcal{X}_2 \times \cdots \times \mathcal{X}_n$ 
denotes the domain of the random variables.  In this paper, we consider only discrete-valued distributions but all our concepts extend to 
differential entropies~\cite{cover2006elements}. The joint entropy $H(S)$ of the variables indexed by $S$ is equal to  
$H(S) =  - \sum_{x_s \in \mathcal{X}_s}  p_s(x_s) \log p_s(x_s) $, where $p_s(.)$ denotes the marginal distribution of the random 
variables belonging to the set $S \subseteq V$. Discrete entropies are known to be non-decreasing submodular set functions--the 
submodularity being a consequence of the data-processing inequality~\cite{cover2006elements}. They are also known to be a strict 
subset of non-decreasing submodular set functions, i.e., when $n> 4 $, there exist set functions which are non-decreasing and submodular 
but not entropies~\cite{Zhang1998}.

The relationship between submodularity and entropies has classically been useful in various probabilistic modeling tasks involving entropies, 
e.g., for proposing approximate algorithms for learning bounded treewidth graphical models~\cite{narasimhan2004pac, Chechetka:2007}, for 
learning naive Bayes models~\cite{krause05near} or for discriminative structure learning~\cite{narasimhanB05}. In this paper, we consider 
transfers in the opposite direction and will extend notions which are usually linked with entropies to all submodular functions. This will 
be achieved through \emph{probabilistic graphical models}.

A joint distribution $p(x)$ on $\mathcal{X}$  is said to factorize in a graph $G=(V,E)$ if and only the distribution $p(x)$ has a factored 
form where each factor depends only on a smaller subset of variables, a clique (for undirected graphs) or a node with its parents (for directed graphs). 
See, e.g.,~\cite{koller2009probabilistic,bishop2006pattern,murphy2012}. In decomposable undirected  and directed graphical models, the entropy 
of the joint distribution decomposes as a sum of marginal entropies of subsets of variables. Moreover, for any distribution, the entropy of the 
closest distribution factorizing in the graphical model provides an  bound on the entropy. For directed graphical models, this last property turns 
out to be a direct consequence of the submodularity of the entropy function. We leverage this property to propose a graphical-model-based upper 
bound for a general class of submodular functions, thus providing a flexible way of defining upper bounds for any submodular function. 
We study these bounds and their properties in detail in \mysec{dag}.

Given a bound $F_G(A)$ on $F(A)$ that depends on a free parameter $G$ (the graph structure), we may bring to bear variational inference techniques~\cite{wainwright2008graphical}: 
we will try to maximize $F_G(A)$ with respect to $A$ while minimizing with respect to the variational parameter $G$. In order to cast this 
variational problem as a convex optimization problem, we will use outer approximations of the marginal polytope~\cite{wainwright2008graphical} 
and inner approximations of the hypertree polytope that represents bounded treewidth graph structures~\cite{narasimhan2004pac,kumar13cvxtjt}. 
We obtain in \mysec{maxsubmod} a saddle point problem which can be solved in polynomial time.

In this paper, we make the following contributions:
\begin{list}{\labelitemi}{\leftmargin=1.1em}
\addtolength{\itemsep}{-.4\baselineskip}
\item[--]
For any directed acyclic graph $G$ and a submodular function $F$, we define in \mysec{dag} a bound $F_G(A)$ and study its properties (monotonicity, tightness). It is specialized to decomposable graphs in \mysec{decomposable}. 
\item[--] In \mysec{maxsubmod}, we propose an algorithm to maximize  submodular functions by maximizing the bound $F_G(A)$ with respect to $A$ while minimizing with respect to the graph $G$, leading to a convex variational method based on outer approximation of the marginal polytope~\cite{wainwright2008graphical} and inner approximation of the hypertree polytope.
\item[--] In \mysec{ext}, we propose extensions to constrained problems and maximizing the difference of submodular functions, which include all possible set functions.
\item[--] We illustrate our  results on small-scale experiments in \mysec{exp}.
\end{list}

{\bf Notations.} 
Throughout this paper, we consider a submodular function $F$ defined on the set $V = \{1, 2, \ldots, n\}$ such that $F(\varnothing)=0$. We use the following definition of submodularity through the diminishing return property:  $\forall A \subseteq B \subseteq V, x \in V \setminus B, F(A \cup \{x\}) - F(A) \geqslant F(B \cup \{x\}) - F(B))$. The main results of the paper do not require additional concepts; in \mysec{ext}, we will need additional concepts such as \lova extensions and base polytopes, which will be presented there. For more details see~\cite{submodtutBach,fujishige2005submodular}.

\section{Directed graphical models}
\label{sec:dag}

In this section, we first review the theory of directed graphical models (for more details, see~\cite{koller2009probabilistic, bishop2006pattern, murphy2012}), and highlight the properties of entropies, which will allow us to define our bounds.

\subsection{Probabilistic directed graphical models}

A joint distribution $p(x)$ on $\mathcal{X} = \mathcal{X}_1 \times \dots \times \mathcal{X}_n$ is said to factorize in  the directed acyclic graph (DAG) $G=(V,E)$ if and only if the distribution $p(x)$  may be written as $p(x) = \prod_{i \in V} p(x_i | x_{\pi_i(G)})$, where $\pi_i(G)$ is the set of parents of node $i$ in $G$. The entropy may then be written as
\BEAS
H(X_V) &  = & \textstyle  - \E_{p(x)} \log p(x) = - \sum_{i \in V} \big\{  \E_{p(x)}  \log p(x_i,x_{\pi_i(G)}) -  \E_{p(x)}  \log p(x_{\pi_i(G)}) \big\} \\
 & = & \textstyle  \sum_{i \in V} \big\{ H(i \cup \pi_i(G)) - H(\pi_i(G)) \big\} .
 \EEAS
When $p(x)$ does not factorize in $G$,  
we define as $p_G(x)$ (and refer to it as the projection of $p$ onto~$G$) the distribution which is closest (in Kullback-Leibler divergence) to $p(x)$ that factorizes in~$G$. Since maximum-likelihood parameter estimation decouples in directed graphical models, a short calculation shows that $p_G(x) = \prod_{i \in V} p(x_i | x_{\pi_i(G)})$ and that the KL-divergence is equal to 
$ D(p|| p_G) =  \sum_{i \in V} \big\{ H(i \cup \pi_i(G)) - H(\pi_i(G)) \big\} - H(V) 
$. Thus, the quantity $H_G(V) \defeq  \sum_{i \in V} \big\{ H(i \cup \pi_i(G)) - H(\pi_i(G))) \big\}$ is always an bound on $H(V)$ and is equal to $H(V)$ if and only if $p(x)$ factorizes in $G$.  

{\bf Marginalization.}
Given a graph $G=(V,E)$, we define by $G_A$ the graph restricted to $A \subseteq V$ i.e.,  $G_A = (A, E \cap ( A \times A))$. In general, if $p$ factorizes in $G$, $p_A$ does not factorize in $G_A$, unless $A$ is an \emph{ancestral set}, i.e., all parents of all elements of $A$ are in $A$ (in other words, we may recursively remove leaf nodes and preserves the factorization). In the following, we denote by $H_G(A)$ the entropy of the projection $p_{G_A}(x_A)$  of $p_A(x_A)$ onto $G_A$. Note that $p_{G_A}(x_A)$  is different in general from $(p_{G})_A(x_A)$ (which is the marginal of the projection of $p$ onto $G$). We have
\BEQ
\label{eq:HG}
H_G(A) = \sum_{i \in A} \big\{ H( A \cap( i \cup \pi_i(G) ) - H(A \cap\pi_i(G)) \big\}.
\EEQ  From properties of entropies and graphical models, we have $H(A) \leqslant H_G(A)$ for any DAG $G$ and set $A \subseteq V$. We  show in the next section that this property  turns out to be a consequence of submodularity.

{\bf Structure learning.} Although we will not use structure learning in this paper, it is worth noting that several entropy-based approaches have been considered for finding the best possible graph (with some constraints) given a probability distribution. They are based on the decomposition of entropies and local search~(see, e.g.,~\cite{Chickering02JMLRb} and references therein).

\subsection{Bounds on submodular functions}

Given a submodular function $F:2^V \to \mathbb{R}$ such that $F(\varnothing)=0$, following \eq{HG}, we   define $F_G$ as
\begin{eqnarray}
F_G(A)
    \label{eq:fg}      & = &   \sum_{i \in A} \Big\{ F\big(A \cap (\pi_i(G) \cup \{i\}) \big) - F\big(A \cap \pi_i(G) \big) \Big\}.     \\[-.25cm]
    \nonumber
\end{eqnarray}
When $F$ is an entropy function, $F_G(A)$ is the entropy of the distribution closest to the distribution of $X_A$ that factorizes in $G_A$ (which is not equal to the marginal entropy on $A$ of the closest distribution that factorizes in $G$). We  now show that $F_G$ bounds $F$ and that the bound is tight for some subsets of $V$ (see all additional proofs in the supplementary material).
\begin{proposition}[Upper bound]
\label{prop:ub}
Let $F$ be a submodular function and $G$ a directed acyclic graph. The function $F_G$ defined in \eq{fg} bounds $F$, i.e., for
all $A \subseteq V$, $F(A) \leqslant F_G(A)$.
\end{proposition}
\textbf{Proof} Without loss of generality, we assume that $\{1, \ldots, n\}$ is the topological ordering (i.e., $j \in \pi_i(G) \Rightarrow i > j$), without loss of generality. For all $A \subseteq V$,
\begin{eqnarray}
\!\! F(A) \!  &   \!\!\!=   \!\!\!& \sum_{i = 1}^n F(A \cap \{1, \ldots, i\}) - F(A \cap \{1, \ldots, i-1\}) \text{ by telescoping the sums}, \nonumber \\
     &  \!\!\! \leqslant  \!\!\! & \sum_{i \in V}   F\big(A \cap (\pi_i(G) \cup \{i\}) \big) - F\big(A \cap \pi_i(G) \big) \text{ by submodularity, since $\pi_i(G) \subset \{1, \ldots, i-1\}$}, \nonumber \\
     &   \!\!\!  =   \!\!\! & F_G(A). \nonumber  \hspace*{11.3cm} \BlackBox   
\end{eqnarray}

\begin{proposition}[Tightness of the bound] 
\label{prop:bound}
For any element, $i \in V$, and any subset $B$ of $\pi_i(G)$, i.e., $B \subseteq \pi_i(G)$,  $F_G( B \cup \{i \}) - F_G(B) = F( B \cup \{i \}) - F(B)$.
\end{proposition}
Note that a corollary of Prop.~\ref{prop:bound} is that the bound is tight on all singletons (by considering  $B = \varnothing$). This implies that any modular properties of $F$ are preserved (and this notably implies that without loss of generality, we may consider only non-decreasing functions).
The bound also has other interesting monotonicity properties, which we now show. 
\begin{proposition}[Monotonicity of bounds - I]
\label{prop:bound1}
If $G'$ is a subgraph of the DAG $G$, then $F_{G'} \geqslant F_G   \geqslant F$,
i.e., for all $A \subseteq V$, $F_{G'}(A) \geqslant F_G(A) \geqslant F(A)$.
\end{proposition}
The following proposition shows that the difference between $F_G(V)$ and $F(V)$ (i.e., approximation for the full set) dominates the error for a specific class of subsets $A$, namely \emph{ancestral sets}. These sets are also the sets $A$ for which $p_A(x_A)$ factorized in $G_A$~\cite{Lauritzen}.
\begin{proposition}[Monotonicity of bounds - II]
\label{prop:bound2}
If $A \subset V$ is an ancestral set of the DAG $G$, then $ 0 \leqslant F_G(A) - F(A) \leqslant F_G(V) - F(V)$.
\end{proposition}

Note that the bound in Prop.~\ref{prop:bound2}, does not hold if $A$ is any subset of $V$. A simple counter-example may be obtained from the entropy of discrete distributions that factorize in the graphical model defined by~$G$: in this case, $F_G(V) = F(V)$, but, for two leaf nodes $\{i,j\}$, $F_G(\{i,j\}) = F_G(\{i\}) + F_G(\{j\}) = F(\{i\}) + F(\{j\})$, which can only be equal to zero (i.e., between zero and $F_G(V) - F(V) = 0$), if the variables indexed by $i$ and $j$ are independent, which is not the case in general if the DAG has a single connected component.
\begin{proposition}[Submodularity]
If the DAG is a directed tree (at most one parent per node), then the bound $F_G(A)$ defines a submodular set function.
\end{proposition}
Finally, when two DAGs are Markov equivalent, the two bounds are equal:
\begin{proposition}[Markov equivalence]
\label{prop:equiv}
If $G = (V,E)$ and $G'= (V,E')$ are two Markov equivalent graphs, then for all $A \subset V$, $F_G(A) = F_{G'}(A)$.
\end{proposition}

\section{Decomposable graphs}
\label{sec:decomposable}

Given an undirected graph $G=(V,E)$, a distribution $p(x)$ is said to factorize in $G$ if $p(x)$ is a product of functions $f_C(x_C)$ that depend only on variables $x_C$, where $C$ is a (maximal) clique. In general undirected models, the entropies do not factorize. We now consider a subclass of graphical models for which the entropy decomposes, namely \emph{decomposable} graphical models. These models may be seen from different views which we now present.

{\bf Triangulated graphs.} A graph $G=(V,E)$ is said to be triangulated if it contains no 
chordless cycles of length greater than 3~\cite{koller2009probabilistic}. 
 A vertex is {\em simplicial} if its neighbours in the graph form
a clique. A graph is {\em recursively simplicial} if it contains a simplicial vertex $i \in V$ and when $i$ is removed, the subgraph that remains is recursively
simplicial.  A  graph is triangulated if and only if it is   recursively simplicial~\cite{Lauritzen}. A perfect elimination ordering is the order in which simplicial vertices can be removed
from the graph. The neighbors of the vertex $i \in V$ that are removed after the vertex~$i$ is eliminated is denoted by $\pi_i(G)$~\cite{golumbic04}. This naturally defines a directed acyclic graph $G$ such that if $p(x)$ factorizes in the graph $G$,  $p(x)$ factorizes in the corresponding DAG, i.e., 
$p(x) = \prod_{i \in V} \frac{p(x_{\pi_i(G) \cup \{i\}})}{p(x_{\pi_i(G)})}$. Hence, decomposable graphical models are a particular case of directed acyclic 
graphs~\cite{Lauritzen}, and thus all properties of directed models shown in \mysec{dag} will be extended to decomposable graphs.  Note that the invariance of our bounds to 
Markov equivalence (Prop.~\ref{prop:equiv}) is key to obtaining a well-defined bound (see Prop.~\ref{prop:bound-dec}).

The most common way to study decomposable graphs is through junction trees, which we now present (algorithmically the simplicial representation is complex to learn graph structures due to its recursive nature).

{\bf Junction trees.}
If $p$ factorizes in $G$, then there exists a 
 junction tree of maximal cliques so that  the joint probability distribution is given by
 $$
p_G(x) = \frac{\prod_{C \in \mathcal{C}{(G)}} p_C(x_C)}{\prod_{(C,D) \in \mathcal{T}{(G)}} p_{C \cap D}(x_{C \cap D})},
$$
where $\mathcal{C}(G)$ denotes the maximal cliques of the graph $G$ and $\mathcal{T}(G)$ denotes the separators represented by the
edges in the corresponding junction tree representation of the graph~\cite{Lauritzen}. Note that these edges are also referred to 
as the {\em minimal separators} of the graph~\cite{golumbic04}. In this paper, we refer to $\mathcal{T}(G)$ as edges in the context of junction trees and minimal separators in the context of set functions. A tree structure $\mathcal{T}(G) \subset  \mathcal{C}(G) \times  \mathcal{C}(G)$ 
may be defined on the set of maximal cliques $\mathcal{C}(G)$, so that (a) neighbors in the clique tree have at least one node in 
common, and (b) the \emph{running intersection property} is satisfied (i.e., the subtree of all cliques containing any given vertex is connected).
The entropy defined on the decomposable graph is then given by $H_G(V) = \sum_{C \in \mathcal{C}(G)} H_G(C) - \sum_{(C,D) \in \mathcal{T}(G)} H_G(C \cap D)$.

\subsection{Bounds on submodular functions}
\label{sec:dec}

We now define the bound of the submodular function $F$ by projection onto a decomposable graph $G = (V, E)$. 
Using  recursive simpliciality, we define
the projection function $F_G$, similar to that of \eq{fg} as:
\begin{equation}
\label{eq:dec1}
F_G(A) = \sum_{i \in V}  \Big\{ F(  A \cap (  \pi_i(G) \cup  \{ i \} ) ) - F(   A \cap \pi_i (G) )  \Big\},
\end{equation}
where $\pi_i(G)$ denotes the neighbors of the simplicial vertex $i$ during its elimination. We also define
an equivalent bound with the junction tree representation; the projection function $F_G$, similar to \eq{fg}, is then given by
\begin{equation}
\label{eq:dec2}
F_G(A) = \sum_{C \in \mathcal{C}(G) } F( C \cap A) - \sum_{(C,D) \in  \mathcal{T}(G)}  F ( C \cap D \cap A) .
\end{equation}
We can now show that the two definitions are equivalent and derive corollaries of Props.~\ref{prop:bound}, \ref{prop:bound1}, \ref{prop:bound2}, for decomposable graphs (see the proof in supplementary material).
\begin{proposition}[Bounds for decomposable graphs]
\label{prop:bound-dec}
Let $F$ be a submodular function. Let $G$ be a decomposable graph. The  set function defined in \eq{dec1} and \eq{dec2} are equal and are bounds on the set function $F$. Moreover,

(a) the bounds are tight on all cliques of the graph $G$,

(b) any decomposable subgraph of $G$ will lead to a looser bound,

(c) if $A$ is obtained by recursively removing simplicial vertices of the graph $G$, then we have $ 0 \leqslant F_G(A) - F(A) \leqslant  F_G(V) - F(V) $.
\end{proposition}

\subsection{Decomposable graph structure learning } 
 \label{sec:lear}
 
We have shown that a submodular function  $F$, when projected onto a decomposable graph $G$, gives an bound $F_G$ with 
interesting monotonic properties. In the next section, we will try to optimize the graph.
Maximum likelihood structure learning happens to be equivalent to minimizing $F_G(V) -F(V)$ with respect to the graph. Typically, the set of decomposable graphs is restricted to have cliques of size $k+1$, which leads to a \emph{treewidth} bounded by $k$ (the treewidth of a decomposable graph is exactly the maximal size of a clique minus one~\cite{Lauritzen}). These graphs are usually considered because inference in these graphs may be performed in polynomial time, with a degree that grows linearly in $k$.

Some properties of maximum likelihood structures   may be transferred to the general submodular case. For example,
the best approximation is always given  by   \emph{maximal junction trees}~\cite{SzantaiK12}, i.e., 
decomposable graphs with maximal cliques of size $k+1$ and separators of size $k$. Therefore, we consider only the space of maximal junction trees with  
treewidth $k$. For these decomposable graphs, denoting $\mathcal{D}_k$ the set of subsets of $V$ with cardinality less than $k+1$, we have
$$F_G(A) = \sum_{ C \in \mathcal{D}_k} \nu_C F( C \cap A) \defeq F_\nu(A)$$
for a certain $\nu \in \rb^{\mathcal{D}_k}$,  with $\nu_C$ being zero for $|C| \leqslant k-1$.  We denote by $\mathcal{J}_k \subset \rb^{\mathcal{D}_k}$ the convex 
hull of all such vectors $\nu$ that correspond to a maximal decomposable graphical model with treewidth equal to $k$. We denote the subsets of $V$ with 
cardinality $k+1$ as $\mathcal{D}_k^{max}$, which we use in \mysec{maxsubmod}.

Given $A \subset V$, the problem of learning the structure of the graph is to minimize $F_\nu(A)$ with respect to $\nu$ in the extreme points of $\mathcal{J}_k$, and since the objective is linear, this is equivalent to optimizing over the entire set $\mathcal{J}_k$. While the problem is NP-hard~\cite{srebro2002maximum}, 
several algorithms have been designed, based on local search techniques~\cite{SzantaiK12}, submodular function minimization~\cite{narasimhan2004pac} or convex relaxations~\cite{kumar13cvxtjt}.

{\bf Special case of trees.}
When $k=1$, maximal decomposable graphs with treewidth equal to $k$ are simple trees, and the problem of finding the best graph is equivalent to a maximum weight spanning tree problem~\cite{Chow68approximatingdiscrete}, which can thus be found in polynomial time.

\section{Variational submodular function maximization}
\label{sec:maxsubmod}

We now show how the bounds described in \mysec{decomposable} may be used for submodular function maximization. Given our graphical model framework, we follow the tree-reweighted framework of~\cite{wainwright2008graphical}.
Given a vertex $\nu$ of $\mathcal{J}_k$ (i.e., the incidence vector of a decomposable graph), we have the bound
$$
\forall A \subset V, \ F(A) \leqslant \sum_{ C \in \mathcal{D}_k  }  \nu_C F(C \cap A) = \sum_{ C \in \mathcal{D}_k  } F(C) \nu_C 1_{C \subset A}.
$$
Since the objective function is linear in $\nu$, for all $A \subset V$, $ F(A) \leqslant  \min_{ \nu \in \mathcal{J}_k}
 \sum_{ C \in \mathcal{D}_k  } F(C) \nu_C 1_{C \subset A}$.
 We may thus  obtain an bound on $\max_{A \subset V} F(A)$ as
\BEAS
\max_{ A \subset V} F(A) & \leqslant & \max_{A \subset V} \min_{ \nu \in \mathcal{J}_k} 
 \sum_{ C \in \mathcal{D}_k  } F(C) \nu_C 1_{C \subset A}.
 \EEAS
 Using weak duality, we obtain:
 \BEAS
\max_{ A \subset V} F(A) & \leqslant &  \min_{ \nu \in \mathcal{J}_k} 
\max_{A \subset V}
 \sum_{  C \in \mathcal{D}_k   } F(C) \nu_C 1_{C \subset A}.
 \EEAS
 
 We may equivalently parameterize $A \subset V$ as $x \in \{0,1\}^n$ through the bijection $A \mapsto 1_A$. This leads to the bound 
\BEAS
\max_{ A \subset V} F(A) & \leqslant &  \min_{ \nu \in \mathcal{J}_k} 
\max_{x \in \{0,1\}^n }
\sum_{ C \in \mathcal{D}_k   }F(C)  \nu_C  \prod_{i \in C} x_i .
\EEAS

The maximization problem $\max_{x \in \{0,1\}^n }
 \sum_{ C \in \mathcal{D}_k   } \nu_C F(C)  \prod_{i \in C} x_i$ is typically NP-hard (however, it is not NP-hard when $\nu$ is an extreme point of $\mathcal{J}_k$). We may relax it by first introducing the set
 $$ \mathcal{M}_k = \Big\{ y \in \{0,1\}^{ \mathcal{D}_k}, \ \exists x \in \{0,1\}^n, y_C=  \prod_{i \in C} x_i \Big\}.$$
 The maximization problem may then be reformulated as
 $\max_{y \in \mathcal{M}_k }
 \sum_{ C \in \mathcal{D}_k   } \nu_C F(C)  y_C$, and thus on the convex hull of $\mathcal{M}_k$. 
  This convex hull is usually referred to as the \emph{marginal polytope}~\cite{wainwright2005new, wainwright2008graphical} and has exponentially many vertices and faces. A common outer relaxation is based on considering only the local consistencies between probabilities defined by $y_C$, $C \in \mathcal{D}$. This leads to~\cite{wainwright2004treewidth}
 $$ \mathcal{N}_k = \bigg\{ y \in [0,1]^{ \mathcal{D}_k}, \forall D \in \mathcal{D}_k^{max},  \forall C \subset D, \ \sum_{ B: C \subseteq B \subseteq D } (-1)^{|B \setminus C|} y_B \geqslant 0 \bigg\}
.$$
We may now state the main proposition of this section:
\begin{proposition}
\label{prop:varMax}
Let $F$  be a submodular function. Then
$$
\max_{A \subset V} F(A) \leqslant  \min_{ \nu \in \mathcal{J}_k} \max_{y \in \mathcal{N}_k}  \sum_{ C \in \mathcal{D}_k   }F(C)  \nu_C y_C
= \max_{y \in \mathcal{N}_k}  \min_{ \nu \in \mathcal{J}_k}  \sum_{ C \in \mathcal{D}_k   }F(C)  \nu_C y_C.
$$
If there exists a $k$-bounded treewidth decomposable graph $G$ such that for all $A \subset V$, $F_G(A) = F(A)$, then the bound is tight.
\end{proposition}

 The last proposition shows that a convex saddle point problem may be considered to provide an bound for $\max_{A \subset V} F(A)$ and that it is tight for certain submodular functions.
Note that the tightness result is still valid if we restrict $\mathcal{J}_k$ to a subclass of graphical models that includes the graph~$G$.
The proof of the previous proposition is  a consequence of the exactness of the relaxation of inference in graphical models, based on outer relaxations of the marginal polytope and its relationship to the Sherali-Adams hierarchy~\cite{wainwright2004treewidth}. By increasing the treewidth $k$, we can get tighter relaxations for growing sets of submodular functions, thus replacing  set functions which are low-order polynomials of the indicator vectors  by submodular functions. Note that these two sets are not included in one another (see also differences of submodular functions in \mysec{ext}).

 \paragraph{Rounding.} Given optimal vectors $y$ and $\nu$, following~\cite{SonGloJaa_optbook}, a set may be obtained by thresholding the values of $y_{\{k\}}$ for all singletons.

\subsection{Optimization algorithm}
\label{sec:varinf}
In this section, we propose an algorithm to optimize the variational bound for maximizing submodular functions in Prop.~\ref{prop:varMax}.
We denote by 
$
\mathcal{P}(\nu, y) =  \sum_{ C \in \mathcal{D}_k   }F(C)  \nu_C y_C 
$ the bilinear cost function, and 
 the goal is to perform the following optimization
\begin{equation}
\min_{ \nu \in \mathcal{J}_k}  \max_{y \in \mathcal{N}_k}  \mathcal{P}(\nu, y),
\label{eq:primal}
\end{equation}
where the two domains are polytopes. 
We are going to use a simplicial method~\cite{bertsekas2011unifying}, which operates as follows.

We denote by $\mathcal{R}(\nu)$ the convex function $\max_{y \in \mathcal{N}_k}  \mathcal{P}(\nu, y)$. Our problem is to minimize $\mathcal{R}(\nu)$ on $\mathcal{J}_k$.
Given a set of extreme points $\nu_1,\dots,\nu_t$ of $\mathcal{J}_k$, we will minimize $\mathcal{R}(\nu)$ not on $\mathcal{J}_k$, but only on the convex hull  $\mathcal{J}_k^{t}$ of all points $\nu_1,\dots,\nu_t$, thus obtaining a point $\bar{\nu}_t$ and the corresponding optimal vector $y_t$ at $\bar{\nu}_t$. This point $\bar{\nu}_t$ is optimal if and only if $\min_{ \nu \in \mathcal{J}_k} \mathcal{P}(\nu, y_t) = \mathcal{P}(\bar{\nu}_t, y_t)$. If the equality above
is met, we have the optimal solution; if  not, then any minimizer $\nu_{t+1}$ of $\min_{ \nu \in \mathcal{J}_k} \mathcal{P}(\nu, y_t)$ may be added to the list of extreme points and the algorithm iterates.

This algorithm converges in finitely many iterations~\cite{bertsekas2011unifying} for polytopes. However, the number of iterations is not known a priori (much like the simplex method). Given the algorithm described above, there are still two algorithmic pieces that are missing: obtaining 
$\min_{ \nu \in \mathcal{J}_k^{t}}  \max_{y \in \mathcal{N}_k}  \mathcal{P}(\nu, y)$, i.e., the optimization problem on the convex hull, and computing 
$\min_{ \nu \in \mathcal{J}_k} \mathcal{P}(\nu, y_t)$, i.e., finding the next graph to add.

{\bf Optimization on the convex hull.}
Since $\mathcal{N}_k$ is defined by polynomially many linear inequalities, we may introduce Lagrange multipliers $z_{CD}$ for each of the constraints 
$\sum_{ B: C \subseteq B \subseteq D } (-1)^{|B \setminus C|} y_B \geqslant 0 $, for $D \in \mathcal{D}_k^{\max}$ and each subset $C$  of $D$. This leads to
\BEAS
\max_{y \in \mathcal{N}_k}  \mathcal{P}(\nu, y)
& = & \min_{ z \geqslant 0 } 
\max_{y \in [0,1]^{\mathcal{D}_k}}   \bigg\{  \sum_{ C \in \mathcal{D}_k   }F(C)  \nu_C y_C + \sum_{(C,D)} z_{CD} \bigg(
\sum_{ B: C \subseteq B \subseteq D } (-1)^{|B \setminus C|} y_B \bigg) \bigg\} \\[-.25cm]
& \defeq & \min_{ z \geqslant 0 }   \mathcal{Q}(\nu, z),
\EEAS
with $ \mathcal{Q}(\nu, z)$ a function which may be computed in closed form (as the maximum of an affine function with respect to $y \in [0,1]^{\mathcal{D}_k}$), and which is jointly convex in $(\nu,z)$. Our optimization problem is then equivalent to
$$
\min_{ \eta \geqslant 0, \eta^\top 1 = 0} \min_{ z \geqslant 0} \mathcal{Q}\Big(\sum_{i=1}^t \eta_i \nu_i, z\Big),
$$
which can be solved by projected subgradient descent techniques, that can obtain both approximate primal variables $(\eta,z)$, but also dual variables $y$~\cite{nedic2009approximate}.

{\bf Finding optimal graphs.}
When $k=1$, maximizing linear functions over $\mathcal{J}_k$ is a maximum-weight spanning tree problem. However, as mentioned in \mysec{lear}, it is NP-hard as soon as $k>1$.  There are two ways of dealing with the impossibility of maximizing linear functions: (a) using a reduced convex hull by generating a large number of random graphs--a strategy often used in variational inference in graphical models, or (b) approximate minimization~\cite{narasimhan2004pac,kumar13cvxtjt}. In this situation, the algorithm still provides an bound on the submodular maximization problems, but the algorithm may stop too early.

\section{Extensions}
\label{sec:ext}
{\bf Difference of submodular functions.}
As shown by~\cite{narasimhanB05}, any set function may be written as the difference of two submodular functions $F$ and $H$. In order to maximize $F(A) - H(A)$, we can use the variational formulation $H(A)  = \max_{s \in B(H)} s^\top 1_A$, where  
$B(H) = \{ y \in \rb^n, \ \forall A \subset V, \ y^\top 1_A \leqslant H(A), \ y^\top 1_V= H(V) \}$ (see, e.g.,~\cite{submodtutBach,fujishige2005submodular}). We then have, for all $A \subset V$, $\nu \in \mathcal{J}_k$ and $s \in B(H)$, $F(A) - H(A) \leqslant F_\nu(A) - s^\top 1_A$. This leads to the convex relaxation:
$$
\max_{A \subset V} F(A)  \leqslant 
\max_{y \in \mathcal{N}_k}  \min_{ \nu \in \mathcal{J}_k , s \in B(H) }  \sum_{ C \in \mathcal{D}_k   } F(C)  \nu_C y_C - \sum_{k \in V} s_k y_{\{k\}}.
$$

{\bf Constrained problems.}
One common practical benefit of having convex relaxations is their flexibility: it is easy to add constraints on the problem. In our variational framework,
any constraints that can be expressed as convex constraints on $y \in \mathcal{M}_k$ may be added. For instance, it includes the cardinality constraint. 

\begin{figure}[t]
\begin{center}
\begin{tabular}{cc}
\includegraphics[width=0.45\textwidth]{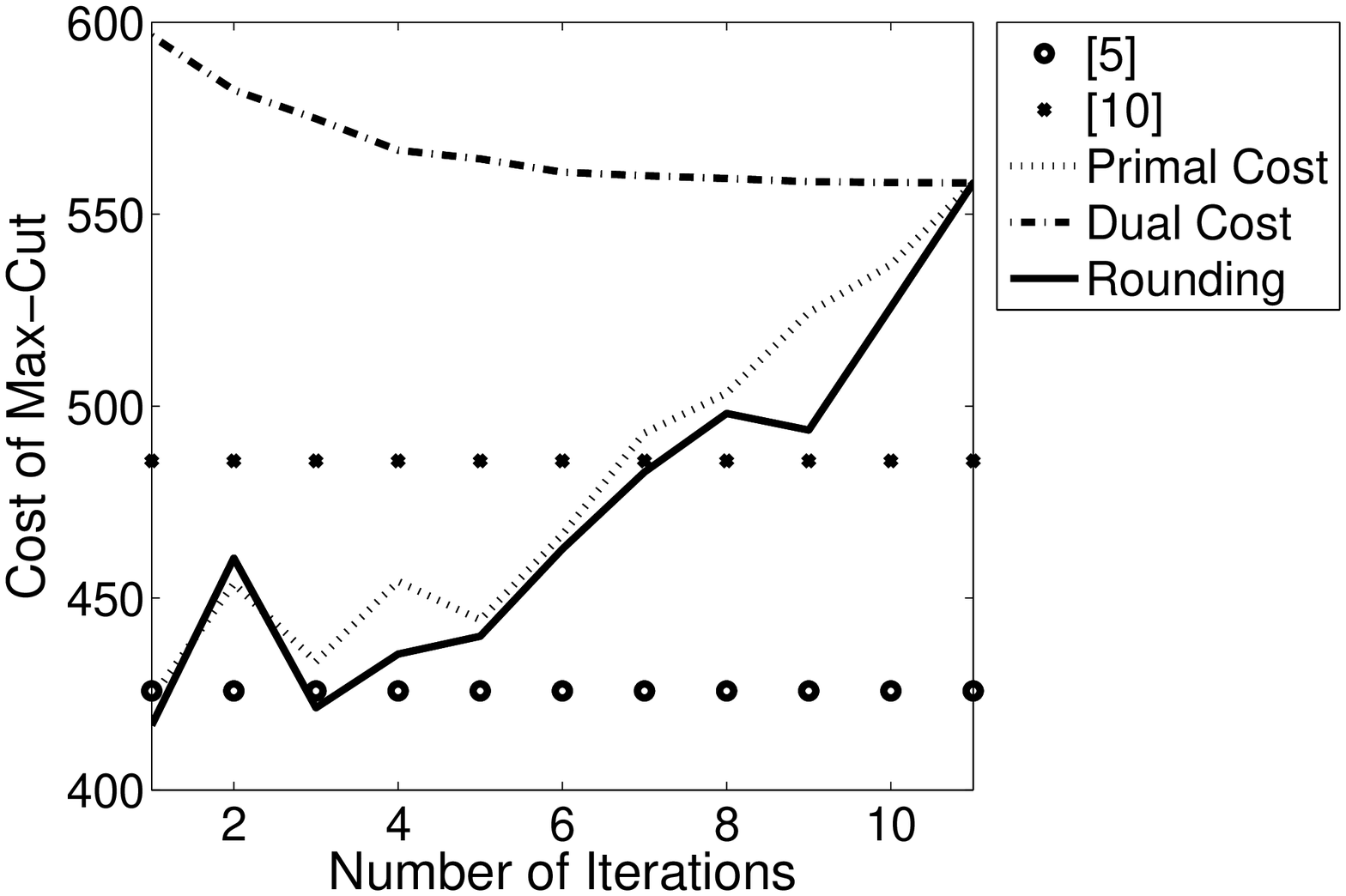}&
\hspace*{-0.5cm}
\includegraphics[width=0.38\textwidth]{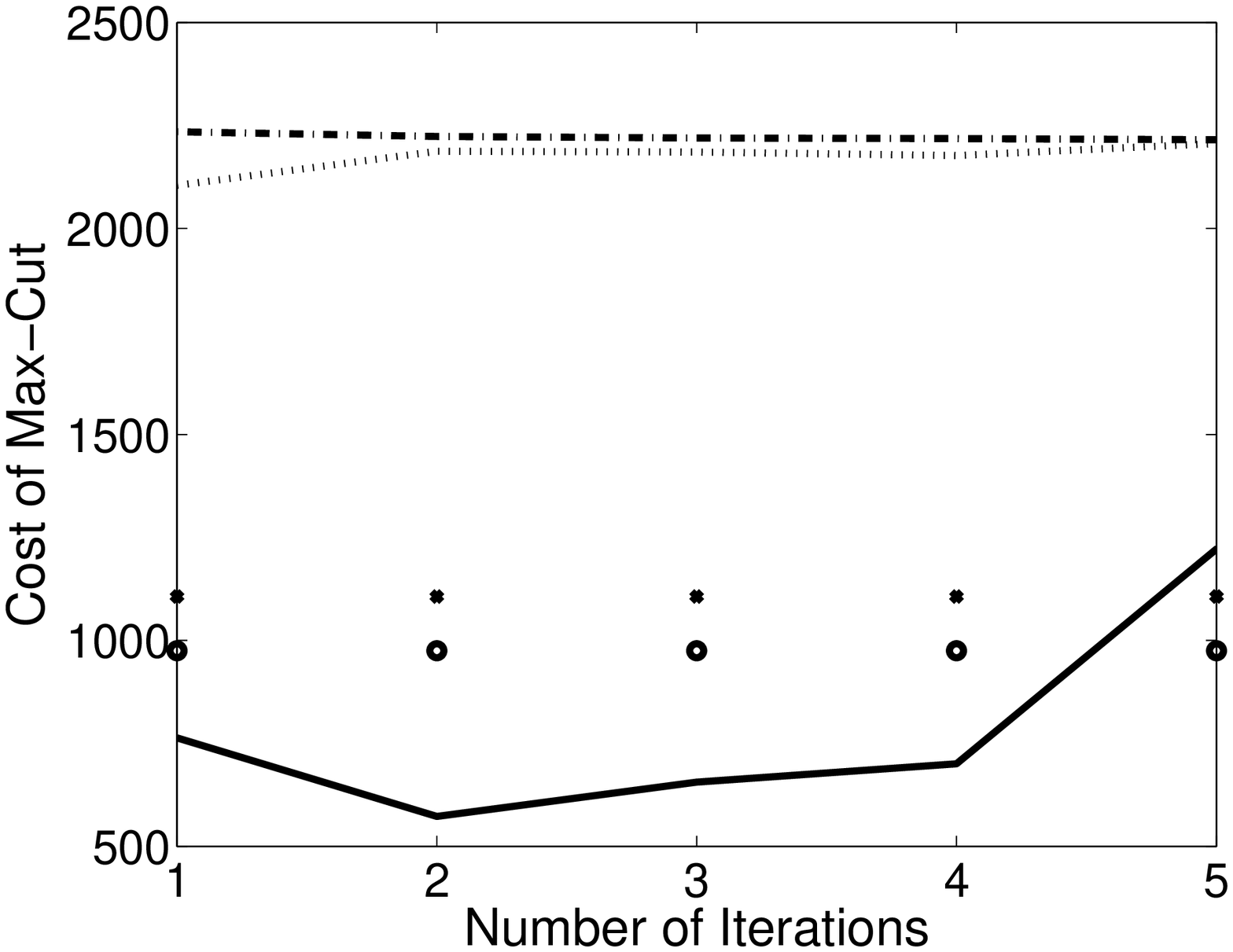}\\
(a) & (b)
\end{tabular}
\end{center}

\caption{Performance on max-cut for (a) 2D-grid and (b) a random graph; the primal cost is $\min_{\nu \in \mathcal{J}_k} \mathcal{P}(\nu, y_t)$ and the dual cost is $\min_{\nu \in \mathcal{J}_k^t} \mathcal{R}(\nu)$ in 
our algorithm. Best seen in color.}
\label{fig:results}
\end{figure}

\section{Experiments}
\label{sec:exp}

In this section, we show the results of our algorithm to solve max-cut on graphs with different configurations: trees, 2D-grid and random graphs. In all our experiments we restrict ourselves to $k=1$, i.e., 
simple spanning trees. Given a set of weights in an undirected graph, $d:V \times V \to \rb_+$, a cut is defined as $F(A) = d(A, V \setminus A) = \sum_{i \in A, j \in V \setminus A} d(i, j)$. The function
$F$ is known to be a non-monotone submodular function. To illustrate our algorithm, we generated synthetic graphs of different configurations with $|V| = 100$ nodes and random positive edge weights. 
In the case of a tree-based cut functions, the algorithm converges to an optimal solution in the first iteration. In the case of 2D grid $(10 \times 10)$, the algorithm converges to an optimal solution 
as shown \myfig{results}-(a). We also show the performance of other constant factor approximation algorithm proposed by Buchbinder et al.~\cite{Buchbinder2012} and Feige et al.~\cite{Feige2011} on 
this configuration. For generating random graphs, we considered $|V| = 100$ nodes with random edge incident on each vertex with probability $0.9$. It can be observed in \myfig{results}-(b)  that our algorithm solves a convex optimization
problem but with a larger integrality gap. This gap could be reduced by using higher treewidth graphs, i.e., $k>1$ instead of trees.

\section{Conclusion}

In this paper, we have developed a novel approximation framework for submodular functions, which enables us to provide convex relaxations of submodular function maximization and related problems. While we have considered only trees in our experiments, it is of clear interest to consider higher treewidths and explore empirically the trade-offs between computational complexity and tightness of our relaxations.

\paragraph{Acknowledgements}

We acknowledge support from the European Research Council grant SIERRA (project 239993). We would also like to thank Nino Shervashidze for detailed feedback on the draft.

\bibliography{smgm}

\appendix

\section{Proof of Proposition 2}
  We have
\BEAS
F_G( B \cup \{i \}) - F_G(B) 
& = &   \sum_{ j \in B \cup \{i \}}  \bigg\{  F(  ( B \cup \{i \} ) \cap (  \pi_j(G) \cup  \{ j \} ) ) - F(   ( B \cup \{i \} ) \cap \pi_j (G) )  \\
& &  \hspace*{3cm} 
- F(  B \cap (  \pi_j(G) \cup  \{ j \} ) ) + F(   B \cap \pi_j (G) ) \bigg\} \\
& = &   \sum_{ j \in B \cup \{i \}}  \bigg\{  F\big[   (B  \cap \pi_j(G)) \cup (\{i\}  \cap \pi_j(G))  \cup \{j\} \big]  - F\big[  (B  \cap \pi_j(G)) \cup (\{i\}  \cap \pi_j(G)) \big]  \\
& &  \hspace*{2cm} 
- F\big[  (B \cap    \pi_j(G)) \cup (B  \cap  \{ j \} ) \big] + F(   B \cap \pi_j (G) ) \bigg\}\\
& = &   \sum_{ j \in B }  \bigg\{  F\big[   (B  \cap \pi_j(G)) \cup \varnothing \cup \{j\} \big] - F\big[  (B  \cap \pi_j(G)) \cup \varnothing  \big]  \\
& &  \hspace*{2cm} - F\big[  (B \cap    \pi_j(G)) \cup    \{ j \}  \big] + F(   B \cap \pi_j (G) ) \bigg\} \\
&  & +    \bigg\{  F\big[   B \cup \varnothing    \cup \{i\} \big]  - F\big[  B \cup \varnothing\big]   - F\big[  B \cup \varnothing \big] + F(   B   ) \bigg\} \\
& = &  F( B \cup \{i \} )  - F(B),
\EEAS
where we have used acyclicity to ensure that for $j \in B$, $\{i\} \cap \pi_j(G)= \varnothing$.

\section{Proof of Proposition 3}

 Let $G = (V, E)$ and $G' = (V, E')$. If $G'$ is a subgraph of $G$, then $E' \subseteq E$ and hence for all the vertices, $i \in V$, $\pi_i(G') \subseteq \pi_i(G)$. Therefore, due to submodularity of $F$,
\begin{eqnarray}
F_G(A)  &  =   & \sum_{i \in A} F(  A \cap (  \pi_i(G ) \cup  \{ i \} ) ) - F(   A \cap \pi_i (G ) ) \nonumber \\
        & \leqslant & \sum_{i \in A} F(  A \cap (  \pi_i(G') \cup  \{ i \} ) ) - F(   A \cap \pi_i (G') ) \nonumber \mbox{ by submodularity}, \\
        &  =   & F_G'(A). \nonumber
\end{eqnarray}

 \section{Proof of Proposition 4}
 Assuming, without loss of generality, that $\{1,\dots,p\}$ is  a topological ordering where $A = \{1,\dots,k\}$, we have
\BEAS
F_G(V) - F(V) & = &   \sum_{i=1}^p \big\{ \big[ F(\{1,\dots,i\}) - F(\{1,\dots,i-1\} \big] - \big[F( \{ i \} \cup  \pi_i (G)) - F(     \pi_i (G)) \big] \big\} \\
& \geqslant &   \sum_{i=1}^k \big\{ \big[ F(\{1,\dots,i\}) - F(\{1,\dots,i-1\} \big] - \big[F( \{ i \} \cup  \pi_i (G)) - F(     \pi_i (G)) \big] \big\} \\
& =& F_G(A) - F(A), 
\EEAS
since all terms are non-negative.

 \section{Proof of Proposition 5}
 For a directed tree  the bound $F_G(A)$ is in fact a quadratic function of the indicator function $1_A$, with quadratic terms equal to $F(\{i,j\}) - F(\{i\}) - F(\{j\})$ which are negative by submodularity of $F$. The function $F_G$ is then a cut function and is submodular.
 
 \section{Proof of Proposition 6}
Two Markov equivalent graphs may be obtained by reversing orders of edges that are not involved in a ``v-structure''. The result is then straightforward.

   \section{Proof of Proposition 7}
   
We first recall the two definitions.

\begin{equation}
\label{eq:dec1}
F_G(A) = \sum_{i \in V}  \bigg\{ F(  A \cap (  \pi_i(G) \cup  \{ i \} ) ) - F(   A \cap \pi_i (G) )  \bigg\},
\end{equation}
\begin{equation}
\label{eq:dec2}
F_G(A) = \sum_{C \in \mathcal{C}(G) } F( C \cap A) - \sum_{(C,D) \in  \mathcal{T}(G)}  F ( C \cap D \cap A) .
\end{equation}

 Equivalence between \eq{dec1} and \eq{dec2}  is a standard result in probabilistic graphical models, which states that if $p(x)$ is a discrete distribution with strictly positive probability mass function that factorizes in $G$, i.e., 
\begin{equation}
p(x) = \frac{\prod_{C \in \mathcal{C}(G) } p_C(x_C)}{\prod_{(C,D) \in  \mathcal{T}(G)}  p_{C \cap D}(x_{ C \cap D})} = \prod_{i \in V} \frac{ p(x_{\pi_i(G) \cup  \{ i \} })}{p(x_{\pi_i (G)} )  }.
\end{equation}

To show tightness of bounds on all cliques, we can always choose an elimination ordering where a given maximal clique is eliminated first, and we then obtain the tightness as a consequence of Prop.~{\bf2}.

In order to show the monotonicity, notice that if $G'$ is a subgraph of $G$, then there is a sequence of decomposable graphs between $G'$ and $G$ so that a single edge is added between two graphs in the sequence~\cite{giudici1999decomposable}. We can then show that at every forward step, the bound has to increase.

Finally, if  a set $A$  is obtained by removing simplicial vertices of the graph, $G$ the relationship between DAGs and decomposable graphs and Prop.~{\bf4} leads to the desired result.

\section{Proof of Proposition 8}

 If $y_C = \prod_{i \in C} x_i$ for all cliques in $\mathcal{D}_k$, then for all $D \in \mathcal{D}_k^{max}$ and $C \subset D$
$$
0 \leq \prod_{i \in C} x_i \prod_{i \in D \setminus C} (1 - x_i) = \sum_{A \subset D \setminus C} (-1)^{|A|} \prod_{A \cup C} x_i,
$$
which imples that $\mathcal{M}_k \subseteq \mathcal{N}_k$.

In the context of probabilistic graphical models, this is equivalent to defining pseudo-marginals $y_C$ on the cliques and ensuring that the pseudo-marginals satisfy the local constraints of the marginal polytope. 
These are the $k^{th}$ order relaxations. The outer relaxation consists of all the extreme points of the marginal polytope as extreme points. However, it also consists of other additional extreme points with 
fractional elements. In the case of decomposable graph models, which are also known as hypertrees, these relaxations are shown to be tight and yield the same optimal solution~\cite{wainwright2004treewidth,wainwright2008graphical}.
Therefore,

\begin{equation}
\max_{A \subset V} F(A) \leq \min_{\nu \in \mathcal{J}_k} \max_{y \in \mathcal{M}} \mathcal{P}( \nu , y) \leq \min_{\nu \in \mathcal{J}_k} \max_{y \in \mathcal{N}} \mathcal{P}(\nu, y)
\label{eq:dir1}
\end{equation}

To prove the tightness, let us assume that there exists a decomposable graph, $G$ denoted by a vertex $\nu_G \in \mathcal{J}_k$ such that $F(A) = F_G(A)$. Therefore,

\begin{eqnarray}
\max_{A \subset V} F(A) & = & \max_{A \subset V} F_G(A)  \nonumber \\
                        & = & \max_{y \in \mathcal{M}} \mathcal{P}(\nu_G, y) \text{ by definition of the marginal polytope}\nonumber \\
                        & = & \max_{y \in \mathcal{N}} \mathcal{P}(\nu_G, y) \text{ as $G$ is a decomposable graph} \nonumber \\
                        & \geq & \max_{y \in \mathcal{N}} \min_{\nu \in \mathcal{J}_k} \mathcal{P}(\nu, y) \label{eq:dir2}
\end{eqnarray}

\eq{dir1} and \eq{dir2} show that they are tight.
 
 \end{document}